# Language Generation for Broad-Coverage, Explainable Cognitive Systems

**Marjorie McShane**                                          MARGEMC34@GMAIL.COM
**Ivan Leon**                                                 LEONI@RPI.EDU
Cognitive Science Department, Rensselaer Polytechnic Institute, Troy, NY, 12180, USA

## Abstract

This paper describes recent progress on natural language generation (NLG) for language-endowed intelligent agents (LEIAs) developed within the OntoAgent cognitive architecture. The approach draws heavily from past work on natural language understanding in this paradigm: it uses the same knowledge bases, theory of computational linguistics, agent architecture, and methodology of developing broad-coverage capabilities over time while still supporting near-term applications.

## 1. Introduction

A core need of artificial intelligence (AI) is human-level language processing, comprised of natural language understanding (NLU) and natural language generation (NLG). The recent successes of large statistical language models such as GPT2/3 and Switch-C (Brown et al., 2020; Fedus, 2021) have ignited widespread interest in NLG as well as the impression that fluent text generation means that the language problem in AI has been solved.[1] But this couldn't be further from the truth due to well-known deficiencies in machine learning-oriented approaches to NLG. For example, such systems don't have any understanding of what they are generating, resulting in errors such as overly repetitive text, word-modeling failures (e.g., writing about fires happening under water), and erratic topic switching (Radford, 2019); they struggle to be contextually relevant since their primary source of context is typically just a dialog history; and they cannot explain their decisions or behavior, leading to distrust of AI systems (Xu, 2019). Moreover, as Bender et al. (2021) argue, there are environmental and ethical considerations to contend with when designing, training, and deploying large language models.

An alternative approach to NLG that avoids these pitfalls uses a knowledge-based approach to cognitive modeling. This paper describes one such program of work, which develops language-endowed intelligent agents (LEIA) within the OntoAgent cognitive architecture (Nirenburg et al., 2021).

The objective of LEIA research is to develop human-level, language-endowed, explainable intelligent systems using content-centric computational cognitive modeling. Language processing in LEIAs follows the theory of Ontological Semantics, which is a human-inspired theory of language understanding that covers both natural language understanding (NLU) and natural language generation (NLG). In the LEIA architecture shown in Figure 1, the left-hand dotted yellow rectangle highlights NLU: an input text is analyzed by the language understanding service, yielding an ontologically-grounded knowledge structure called a text meaning representation, or TMR. We refer to these as $_{NLU}$TMRs to emphasize their source. Correspondingly, the right-hand dotted yellow rectangle highlights NLG: the Attention and Reasoning service creates an ontologically-grounded text meaning representation that serves as a content specification for the process of NLG, which is carried out by the Verbal Action Generator. We call this variety of text meaning representations $_{NLG}$TMRs, i.e., meaning representations that are used as input to NLG.

---

[1] For surveys see Gatt and Krahmer (2018) and Santhanam and Shaikh (2019). For an approach that makes reference to knowledge but is not knowledge-based and explainable in the sense we pursue here, see Dinan et al. (2018).

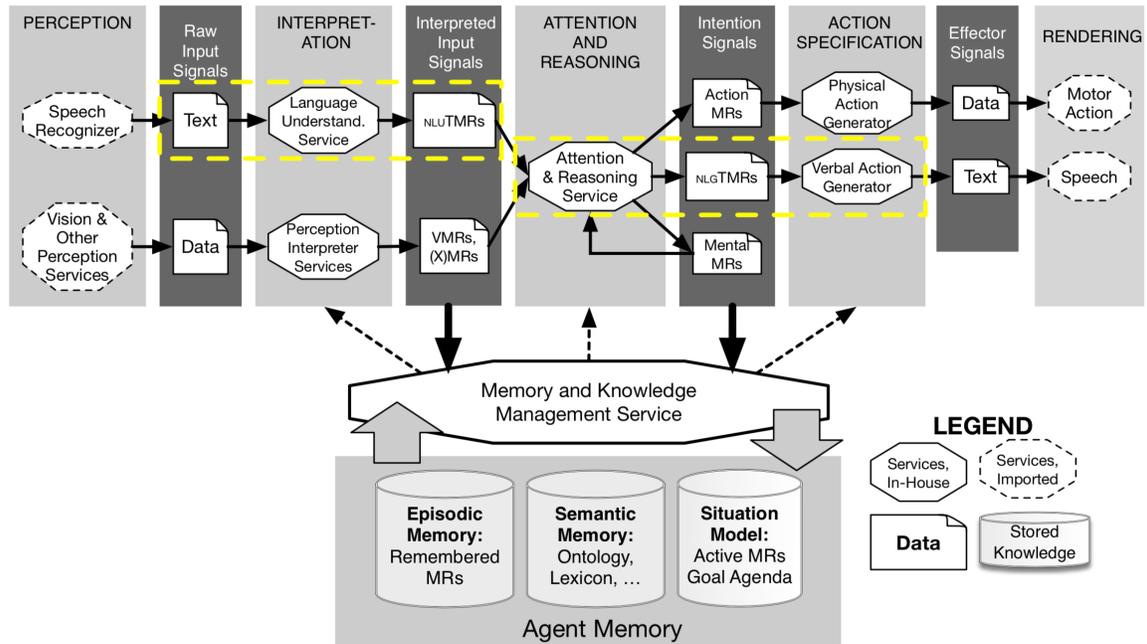

*Figure 1.* The OntoAgent architecture for developing LEIAs.

Until recently, LEIA research has devoted significantly more attention to NLU than NLG (McShane & Nirenburg, 2021). That is, whereas we have pursued broad and deep coverage of language in NLU, we have configured only the minimal NLG capabilities necessary to support the functioning of our prototype agent applications. This decision reflected prioritization in the face of resource constraints. However, it is equally important to approach NLG with the same view toward broad coverage and domain independence. The reason is that small-domain NLG capabilities will not scale up since they involve significant simplifications (*ibid.*).

This paper reports early progress in developing NLG capabilities of LEIAs. The main claims of the paper are: (1) The same knowledge environment that supports NLU can support NLG, offering important economy in knowledge engineering across agent functionalities. (2) Broad-coverage NLG capabilities can be developed and tested using NLU as a source of content specification. (3) NLG capabilities can be developed over time such that early results can contribute over time to more sophisticated capabilities.

## 2. From NLU to NLG

A challenging question in the history of work on NLG has been, What *meaning* does the system want to express through language and where does it come from? This task is known as content specification. In cognitive systems that are developed using typical cognitive architectures (ours included), content specification is an aspect of agent reasoning – something we will return to in a moment. However, it is important to understand, by way of orientation, that most current dialog systems are not cognitive systems – they avoid the task of content specification entirely. Instead, they use ML methods – based on surfacy (not interpreted) text features – to predict the most likely response based on the previous dialog turn. Recent attempts to optimize this approach include Brown et al.'s (2020) work on few-shot learning using GPT-3's language model, and Liu et al's (2020) incorporation of aspects of "persona" (text strings about the speaker) into the content of generation.

Returning to LEIAs, before an agent says anything it must (a) know what content it wants to express and (b) decide to express it using language rather than some other action (e.g., agreement and disagreement can be expressed by nodding and shaking one's head, respectively). This reasoning is carried out by the Attention & Reasoning service (cf. Figure 1), yielding a $_{NLG}$TMR. Figure 2 shows the $_{NLG}$TMR for the meaning that some person attached a painting to a wall on 05.01.2021 at 9:02. The $_{NLG}$TMR is comprised of instances of ontological concepts, distinguished by numerical indices, that are stored in two modules of agent memory: long-term episodic memory and the situation model. In other words, this is a particular FASTEN event (FASTEN-18) involving a particular HUMAN (HUMAN-104), PICTURE (PICTURE-7) and WALL (WALL-40).

| FASTEN-18 | | | HUMAN-104 | |
|---|---|---|---|---|
| AGENT | HUMAN-104 | | AGENT-OF | FASTEN-18 |
| THEME | PICTURE-7 | | PICTURE-7 | |
| DESTINATION | WALL-40 | | THEME-OF | FASTEN-18 |
| DATE | 05.01.2021 | | WALL-40 | |
| CLOCK-TIME | 09:02 | | DESTINATION-OF | FASTEN-18 |

*Figure 2*. An example of an $_{NLG}$TMR – the meaning an agent wants to express using language.

The agent's knowledge about the *types* of concepts used in $_{NLG}$TMRs is stored in the ontology. For example, the semantic constraints on the case-role fillers of FASTEN are shown in the left-hand column of Figure 3. The *default* facet holds more typical fillers than the basic semantic constraints indicated by the *sem* facet.

| Ontology excerpt | | | Episodic memory excerpt | |
|---|---|---|---|---|
| FASTEN | | | HUMAN-104 | |
| AGENT | sem | HUMAN | HAS-NAME | Tom |
| THEME | sem | PHYSICAL-OBJECT | GENDER | male |
| DESTINATION | default | PLACE | AGENT-OF | *a list of all known event instances,* |
| | sem | PHYSICAL-OBJECT | | *including* FASTEN-*18* |

*Figure 3*. Excerpts from the ontology and episodic memory.

The agent's knowledge about concept *instances,* like those that populate ₙₗ₉TMRs, is stored in episodic memory. For example, the fact that this person's name is Tom is listed in the episodic memory frame for HUMAN-104, which includes everything else the agent knows about this person, including his gender, all the event instances he was an agent of in the past, etc. – as shown in the right-hand column of Figure 3.

To generate sentences from ₙₗ₉TMRs, the agent must make a lot of linguistic decisions, such as which of the available synonyms to use in lexical selection, when to use multiword expressions, which morphological forms of words to generate, how to render referring expressions, which voice (active vs. passive) to use for each clause, and so on. The complexity of each of these problems is well attested in the descriptive, theoretical, and computational linguistic literature of the past half century. Faced with this complexity, cognitive systems developers have typically addressed only the subset of issues that are required by the given application, using methods appropriate to that application. This makes sense in the short term, especially to foster progress on integrated agent systems; but it won't scale up. What we are trying to do, by contrast, is to work on issues of NLG in principle, across domains, in ways that are practical and mindful of the current state of the art.

In order to work on a broad range of issues in NLG across domains, it is desirable to accumulate a large inventory and variety of ₙₗ₉TMRs to serve as examples of content specification. The problem is, extant LEIA-based application systems – like most cognitive system applications – do not have broad coverage, thus limiting the availability of ₙₗ₉TMRs generated by the agent's content specification module. However, that turns out to not be an insurmountable hurdle because LEIAs provide us with a workaround: we can use the LEIA's own NLU module to generate knowledge structures that are very similar to ₙₗ₉TMRs and use them as content specifications for NLG. Figure 4 shows how this works. (1) Take any natural language text as input, (2) use the agent's NLU module to generate ₙₗᵤTMRs for them, (3) strip the latter of the metadata indicating which words and constructions were used in the original sentence, and (4) treat the stripped meaning representation *as if* it had been generated by the agent's content generation module. We must emphasize that, while this is *an R&D strategy,* not an application configuration, it offers ample fodder for experimentation without waiting for agent systems to become far more robust than they currently are or resorting to having people manually create meaning representations to serve as content specifications. Of course, the NLG capabilities developed in this way can be used in LEIA-based application systems as well.

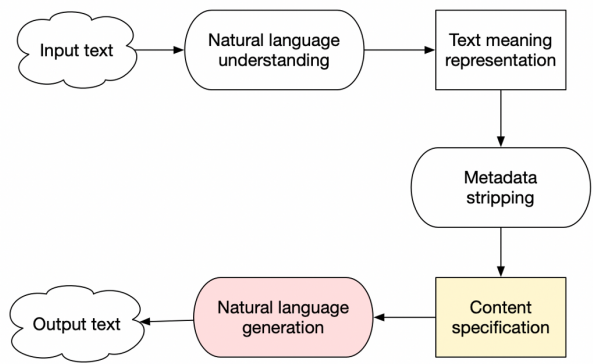

*Figure 4.* NLU provides content specifications to NLG.

Since ₙₗᵤTMRs are important to our story, let us briefly describe their creation – which is a good introduction to language processing by LEIAs overall.

As shown in Figure 1, language processing by LEIAs relies on agent memory, comprised of various static and dynamically populated knowledge elements. The three most important ones for this discussion are: (1) a lexicon containing around 30,000 linked syntactic and semantic descriptions of lexemes, multiword expressions, and constructions; it is far from comprehensive but is sufficient to allow us to work on the core issue of lexical ambiguity; (2) a property-rich ontology containing around 9,000 concepts – each described by dozens of properties – that grounds all semantic descriptions in the lexicon and episodic memory; and (3) a dynamically populated episodic memory containing both the situation model and the interaction history.

Let us walk through the process of NLU using the sentence *Tom secured a paining to the wall* – which is one rendering of the situation illustrated by the ₙₗGTMR in Figure 2. Its ₙₗᵤTMR is shown in Figure 5.

| | | | | |
|---|---|---|---|---|
| FASTEN-1 | | | PICTURE-1 | |
| AGENT | HUMAN-1 | | THEME-OF | FASTEN-1 |
| THEME | PICTURE-1 | | *from-sense* | *painting-n1* |
| DESTINATION | WALL-1 | | *word-num* | *3* |
| TIME | (< find-anchor-time) | | | |
| *from-sense* | *fix-v2* | | WALL-2 | |
| *word-num* | *1* | | DESTINATION-OF | FASTEN-1 |
| | | | COREFER | WALL-1 |
| HUMAN-1 | | | *from-sense* | *wall-n1* |
| HAS-NAME | 'Tom' | | *word-num* | *6* |
| AGENT-OF | FASTEN-1 | | | |
| *from-sense* | *Tom-n1 ; from onomasticon* | | | |
| *word-num* | *0* | | | |

*Figure 5.* The ₙₗᵤTMR for Tom secured a painting to the wall.

To create a ₙₗᵤTMR from this input, the agent carries out preprocessing, syntactic analysis, lexical lookup, semantic analysis, and pragmatic analysis – as described in McShane and Nirenburg (2021). For this abbreviated description of NLU, the first point of interest is the lexicon. Table 1 shows a simplified version of the lexical sense of *fix* that was used to generate the ₙₗᵤTMR.

*Table 1.* Simplified lexical sense for fix-v2.

| Top-level info | | Syn-Struc | | Sem-Struc | |
|---|---|---|---|---|---|
| fix-v2 | | subj | $var1 | FASTEN | |
| def | to attach something | v | $var0 | AGENT | ^$var1 |
| | to something else | directobject | $var2 | THEME | ^$var2 |
| ex | He fixed the bookshelf to the | pp | | DESTINATION | ^$var4 |
| | wall. | prep | $var3 (root to) | ^$var3 (null-sem+) | |
| synonyms | attach, fasten, secure | n | $var4 | | |

This sense is headed by the verb *fix* but covers the synonyms *attach, fasten,* and *secure* as well. Syntactically (see the syntactic structure zone, or syn-struc), it is transitive with an obligatory prepositional phrase headed by 'to'. Semantically (see the semantic structure zone, or sem-struc), it instantiates the concept FASTEN with the AGENT, THEME, and DESTINATION slots filled by the semantic interpretations (indicated by ^) of the coreferenced variables. The ontological description for FASTEN, whose excerpt was introduced earlier, indicates the semantic constraints on the fillers of those case-role slots. When generating a $_{NLU}$TMR, the agent looks up all of the sentence's words in the lexicon and optimizes how they can be combined into a context-sensitive meaning representation. Notice how similar the $_{NLU}$TMR is to the $_{NLG}$TMR in Figure 2 if one strips the metadata shown in italics.

A couple of things to notice are: (a) whereas *a picture* does not have a coreferent, *the wall* does: it is resolved to a previous instance, WALL-1; (b) the filler of the TIME slot, used to reflect the past tense of the verb, is a call to a procedural semantic routine that attempts to ground the utterance in an actual time and then make the time of the reported action before that anchor time; (c) in the $_{NLU}$TMR, Tom's name is actually mentioned, whereas in the $_{NLG}$TMR shown earlier it is not – an issue we will return to later on. Our goals in describing the NLU process here are modest: first, to introduce the kinds of knowledge bases that used for both NLU and NLG; and second, to emphasize that, once we strip the metadata, the results of NLU are very similar to the $_{NLG}$TMR resulting from the agent's reasoning, thus providing a back door into creating content specifications as input to work on broad-coverage NLG.

## 3. The Architecture of NLG

Figure 6 shows the architecture of the NLG module (the Verbal Action Generator module in Figure 1). The subsections below discuss the first four stages: Lexical selection, Converting CandidateSets into CandidateSolutions, Generating candidate sentences, and Selecting the best sentence.

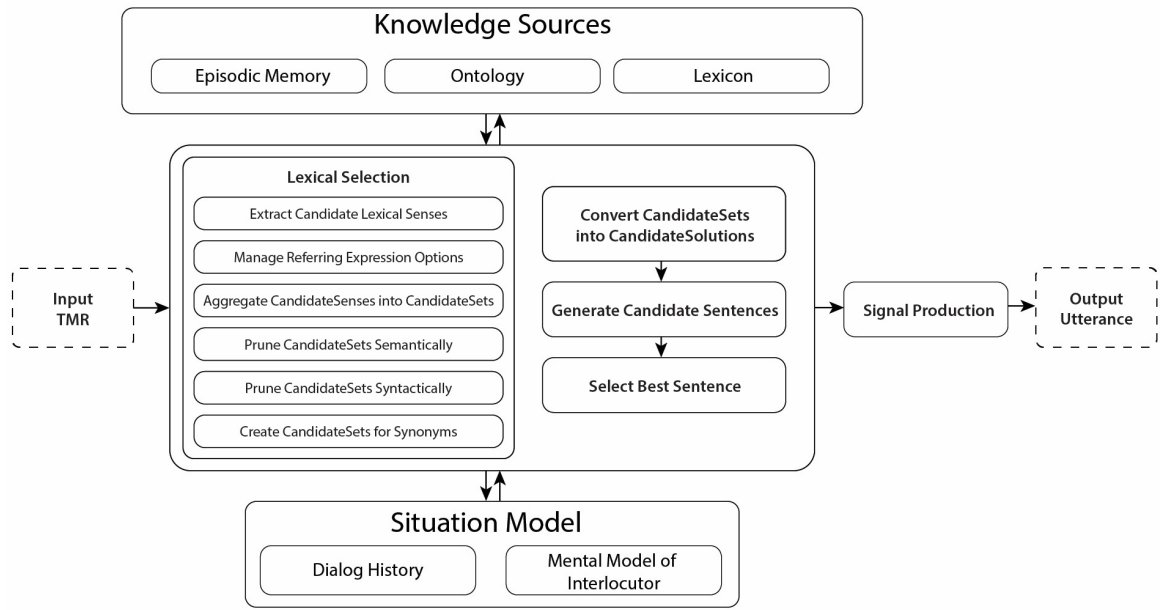

*Figure 6.* The architecture of NLG.

### 3.1 Lexical Selection

Lexical selection is the process of collecting the lexical senses that might ultimately contribute to the final solution for generating a sentence from a <sub>NLG</sub>TMR. It includes five steps that we describe using the example of Tom fastening a painting to the wall.

#### 3.1.1 Extract Candidate Lexical Senses

This step covers two cases: meanings that are recorded in the heads of <sub>NLG</sub>TMR frames, and meanings that are not (typically, property values).

*a. Treat <sub>NLG</sub>TMR heads.* For each frame in the <sub>NLG</sub>TMR, the agent identifies its head concept and searches the lexicon for all CandidateSenses whose sem-struc zones are headed by that concept. Table 2 gives a thumbnail sketch of some of lexical senses headed by the concept FASTEN.

*Table 2.* A subset of lexical senses headed by FASTEN.

| Sense # | Example | Syn-Struc | Sem-Struc |
|---|---|---|---|
| fix-v2 (synonyms: attach fasten, secure) | They fixed the shelves to the wall | S V DO PP-"to" | FASTEN<br>  AGENT *(SEM HUMAN)*<br>  THEME *(SEM PHYSICAL-OBJECT)*<br>  DESTINATION (*SEM PHYSICAL-OBJECT*) |
| affix-v1 | They affixed wallpaper to the wall | S V DO PP-"to" | FASTEN<br>  AGENT *(SEM HUMAN)*<br>  THEME *(SEM PHYSICAL-OBJECT)*<br>  DESTINATION (*SEM PHYSICAL-OBJECT*) |
| moor-v1 | They moored the ship | S V DO | FASTEN<br>  AGENT *(SEM HUMAN)*<br>  THEME **(SEM SURFACE-WATER-VEHICLE)**<br>  INSTRUMENT **(SEM ANCHOR)** |
| skewer-v1 | He skewered the meat | S V DO | FASTEN<br>  AGENT *(SEM HUMAN)*<br>  THEME *(PHYSICAL-OBJECT)*<br>  INSTRUMENT **(SEM SKEWER)** |

In some cases, the sem-struc zone of a lexical sense expresses a meaning that is more precise than the closest available concept in the ontology. This is done by more narrowly constraining case-role fillers and/or including additional property-based descriptions. For example, the meaning of 'moor' is described in the lexicon as a type of FASTEN rather than creating a special concept for MOORING. (Within our knowledge environment these are compatible knowledge acquisition strategies; the choice between them is discussed in McShane & Nirenburg, 2021). In the table, the semantic constraints that are specified in the lexical sense – thus overriding ontological defaults – are in **boldface**. Those that are understood from the ontology are in *italics*. In some cases, synonyms are listed in the synonyms field of a given sense (as in row 1) and in other cases they are listed separately; e.g., affix-v1 could have been listed as a synonym of fix-v2. It is inevitable that large knowledge bases will have some such inconsistencies and they are actually not bugs but reflections of tradeoffs: whereas it is fast to type some extra words into a synonyms field, that makes it harder to keep track of the full inventory of senses of polysemous words when they are separated across the word's head entry and the synonyms fields of other words.

Similarly, the lexicon has many senses of PAINTING, including those in Table 3.

*Table 3.* A subset of senses headed by PAINTING.

| Sense # | Example | Syn-Struc | Sem-Struc |
|---------|---------|-----------|-----------|
| painting-n1 | Let's buy a painting. | N | PAINTING |
| picture-n1 | I like that picture. | N | PAINTING |
| cityscape-n1 | I want to buy a cityscape. | N | PAINTING<br>DEPICTS CITY |
| graffiti-n1 | He hates graffiti. | N | PAINTING<br>LOCATION EXTERIOR-BUILDING-PART<br>LEGALITY-ATTRIBUTE NO |
| landscape-n1 | What a pretty landscape! | N | PAINTING<br>DEPICTS COUNTRYSIDE |

*b. Treat property values.* In most cases, properties do not head NLGTMR frames; instead, they are used as descriptors within OBJECT and EVENT frames. For example, *pretty picture* will be expressed in a NLGTMR as "PAINTING (AESTHETIC-ATTRIBUTE .8)". For all properties (attributes and relations) to be generated in text, the agent must identify all words in the lexicon that can render the given meaning. For our example, *pretty, lovely,* and *attractive* are among the options for expression "AESTHETIC-ATTRIBUTE .8". Suffice it to say that there are many details to be managed with respect to this process that must be addressed both algorithmically (to foresee and account for realization options) and in terms of software engineering.

### 3.1.2 Manage referring expression options

The goal of this stage is to enhance the inventory and description of CandidateSenses by analyzing coreference relations and their effect on the choice of referring expressions. Let us begin with some high-level examples.

The speaker and hearer, like all humans, will be referred to in NLGTMRs using frames headed by HUMAN. This means that, during Step 1, all lexical senses headed by HUMAN (e.g., person-n1, individual-n2, he-n1, she-n1, etc.) will in the set of CandidateSenses. When a NLGTMR indicates coreference with the speaker or hearer, then the overwhelmingly correct decision is to render it as the appropriate pronoun: *I (me, myself)* or *you (yourself).* At this stage, all senses other than *I (me, myself)* or *you (yourself)*, as applicable, are removed from the CandidateSenses.

3rd person HUMANs who are already in the discourse context (situation model) can be referred to using pronouns (*they, them, themselves,* etc.), definite descriptions (*the waiters*), relative referring expressions (*my neighbor*), or various naming conventions (*Bruce and Charlie; the Jones brothers*). Pronouns will already be among CandidateSenses and, at this stage, those with incorrect features (e.g., singular instead of plural) will be pruned out. Full noun phrases that can be used to refer to 3rd person HUMANs may or may not already be in the candidate space – an issue that manifests differently depending on whether we are practicing NLG using metadata-stripped NLUTMRs or carrying out NLG within an application system. If we are practicing NLG using NLUTMRs, then the availability of particular definite descriptions will often be provided explicitly:

e.g., if a frame is headed by WAITER and is described as "HUMOR-ATTRIBUTE .8" and "COREF WAITER-2", then the NLG system knows to generate 'the funny waiter'. By contrast, if the NLG system is working from a content specification created within an application, then individuals will be referred to using their HUMAN anchors in episodic memory: e.g., HUMAN-2002. The agent will not magically know the best way – of all the options listed above – to refer to this individual in this context. This is a complex and well-known problem that spans the tasks of content specification and NLG. It is at the top of our agenda once we have the basic throughput of NLG, described below, solidly in place.

Our final example of reference issues involves non-human OBJECTs, which may or may not already be in the discourse context. If they *are*, then they will be referred to using either a pronoun or a definite description (a noun phrase with *the*). If they are not, then if they are singular they will be referred to using an indefinite description (a noun phrase with *a/an*) and if they are plural they will be referred to using either *some* or no article. Here, again, we encounter more challenges within an end application than when we practice NLG using $_{NLU}$TMRs. When we use $_{NLU}$TMRs, the specificity of the referents will have already been selected (e.g., BANK vs. BUILDING vs. THING), whereas within an end application the agent's content specification and NLG modules must determine the best level of specificity.

In sum, during this stage the agent excludes some lexical senses (e.g., those headed by HUMAN but with the wrong value of number) and decorates others with coreference information about how they should be realized (e.g., using *a/an* or *the*).

### 3.1.3 Aggregate CandidateSenses into CandidateSets

This function generates all permutations of the lexical senses in purview, selecting one sense per meaning for each CandidateSet. For our example, the combination of senses used in the $_{NLU}$TMR in Figure 5 is attach-v2, painting-n1, Tom-n1, wall-n1. As applicable, these senses carry along the reference-related information computed in Step 2 – e.g., the need for the article *a/an* or *the*.

### 3.1.4 Prune CandidateSets Semantically

This step removes CandidateSets that will not work semantically, of which there might be many because the initial lexical retrieval stage cast a very wide net (it looked only at the heads of the sem-struc of lexical senses, not any other aspects of their semantic descriptions or their syntactic descriptions).

Before turning to the details of this step, let us consider the basic idea using the example of requests for action (i.e., commands), which are realized as the concept REQUEST-ACTION. They are ontologically defined as having, among others, the features POLITENESS and REFUSAL-OPPORTUNITY (i.e., the degree to which the speaker leaves the door open to say no), both measured on the abstract scale $\{0,1\}$. Any content-specifying TMR frame headed by REQUEST-ACTION can include values for either or both of these. Similarly, all constructions in the lexicon that are headed by REQUEST-ACTION (there are many dozens) are specified for these values. Some examples:

| POLITENESS 1, REFUSAL-OPPORTUNITY 1 | It might be a good idea for you to VP |
| | It would be much appreciated if you would VP |
| POLITENESS .8, REFUSAL-OPPORTUNITY .8 | Could you please VP? |

POLITENESS 0, REFUSAL-OPPORTUNITY 0     VP<sub>IMPERATIVE</sub>, dammit!

One could quibble over the numeric values of these properties assigned to different language constructions, but the idea should be clear: when the agent generates a content specification (a <sub>NLG</sub>TMR), it does so with knowledge of the situation and its interlocutor. This means that feature-value assertion is a part of the content specification. The NLG engine can then use those feature values to guide its selection of which language utterances are appropriate.

Turning now to details, this step is divided into functions covering lexical senses that do not contain dependencies and those that do contain them.

*a. Evaluate non-argument-taking CandidateSenses***.** For each CandidateSense whose syn-struc does *not* contain dependencies (e.g., basic nouns), score the goodness of the *semantic* match between the given <sub>NLG</sub>TMR frame and the CandidateSense's requirements, listed in the sem-struc. (i) Exclude impossible matches, such as CandidateSetss that use CandidateSenses for which the sem-struc incudes elements not reflected in the <sub>NLG</sub>TMR. For our sample sentence, the agent can immediately reject *cityscape-n1, graffiti-n1,* and *landscape-n1* as candidate realizations because they include semantic descriptions (values of DEPICTS and LOCATION) that are not in the <sub>NLG</sub>TMR. That is, there is no reason to generate a sentence using *landscape* unless we know that the picture in question actually is a landscape. (ii) Give a bonus to exact matches. For example, if our <sub>NLG</sub>TMR had included PICTURE (DEPICTS LANDSCAPE) then CandidateSetss using *landscape-n1* would receive a large bonus. (iii) Additional scoring conditions are under development.

*b. Evaluate argument-taking CandidateSenses.* For the remaining CandidateSenses (i.e., those that *do* contain syn-struc dependencies), score the goodness of the match between the given <sub>NLG</sub>TMR frame and the CandidateSense's semantic description. (i) Exclude impossible matches. For our sample sentence, moor-v1 is excluded because it requires its THEME to be SURFACE-WATER-VEHICLE whereas the TMR's THEME is PAINTING. Similarly, *skewer-v1* is excluded because its explanatory component "INSTRUMENT SKEWER" is not found in the TMR. (ii) Give a bonus to CandidateSets that use CandidateSenses that fulfill more narrow selectional constraints. For example, say our input was actually "They moored the ship": Although SHIP fulfills the THEME constraints of fix-v2 and affix-v1 (by virtue of being a PHYSICAL-OBJECT), the sense moor-v1 gets a large bonus because SHIP matches the much more narrow constraints on its THEME (namely, it's a SURFACE-WATER-VEHICLE); (iii) Additional scoring conditions are under development.

### 3.1.5 Prune CandidateSets Syntactically

For the CandidateSenses still under consideration, determine if all obligatory syntactic components can be filled by <sub>NLG</sub>TMR elements. If they cannot, then the given sense cannot be used to generate a sentence. To understand this, consider the minimal pair of examples "I am walking" (intransitive) vs. "I am walking the dog" (transitive). The lexicon has different senses of *walk* covering each of these constructions. If a <sub>NLG</sub>TMR expresses the "I am walking" situation, then clearly the "I am walking the dog" sense will not fit, and vice versa. Similarly, pronominal realizations cannot be used for OBJECTs that are modified (one cannot say '*blue it'*) or for EVENTs that are modified or have case-roles listed in their frames.[2] In short, lexical senses are *combinations* of syntactic and semantic descriptions and must be selected or rejected as a whole.

### 3.1.6 Create CandidateSets for Synonyms

For the CandidateSenses still under consideration, if synonyms are listed in the synonyms field, then create a set of CandidateSets for each synonym that mirrors that of the sense's head word.

## 3.2 Convert CandidateSets into CandidateSolutions

The CandidateSets the agent has created so far are only sets of lexical senses – in some cases, decorated with reference information. More work needs to be done to prepare them to serve as input to the engine that will generate actual sentences, a process called rendering. For this, we are currently using a software package called SimpleNLG (Gatt & Reiter, 2009). Specifically, for each CandidateSet, the agent (a) expands all argument-taking syn-strucs, filling their variable slots with actual CandidateSenses, (b) orders the argument-taking syn-strucs according their ordering in the NLGTMR, (c) extracts the ordered lists of words from the filled syn-strucs, along with any features they carry, and (d) asserts other features that the SimpleNLG engine can consume.

Without delving too deeply into the particulars of SimpleNLG, some of the features and values that the agent can provide to SimpleNLG are: the set of words that will comprise the sentence and their preliminary ordering (which can be overridden, e.g., if the verbal feature 'passive' is selected); their parts of speech; their syntactic function (e.g., for nouns – subject, direct object, etc.; for verbs – main verb, auxiliary verb); features of verbs (e.g., form – infinitive, imperative, infinitive, etc.; tense – present, past, future); features of nouns (e.g., number – singular, plural; possessive – true, false); features of clauses (e.g., interrogative type – yes/no, how; passive vs. active voice); and components of to-be sentential constituents (e.g., a prepositional phrase needs to be formed from the preposition 'to' and the noun phrase 'a' 'nice' 'cat').

Consider the process of converting CandidateSets into CandidateSolutions for our example of Tom securing a painting to the wall. One of the many CandidateSets includes the words 'secure' (a synonym of attach-v2), 'Tom' (Tom-n1), 'painting' (painting-n1), and 'wall' (wall-n1). The syn-struc of attach-v2 provides the basic word order for the clause: subject (*Tom*), verb (*secured*), direct object (*painting*), prepositional phrase (*wall*). It also indicates that the word 'to' must be used as the preposition that introduces the noun phrase indicating the DESTINATION (*wall*). To emphasize, 'to' was not in the original list of words – it got added by the syn-struc of attach-v2. From reference processing, we know that *painting* needs the article 'a' whereas *wall* needs the 'the'; these articles need to be added, in the appropriate order, to the set of words that will be passed to the sentence generator. Finally, from the NLGTMR we know that this happened in the past, so the verb requires the feature past tense, which will be realized appropriately by SimpleNLG. All of this information is compiled into the input format required by the SimpleNLG engine that will be launched next. Each such data structure that will serve as input to SimpleNLG is what we call a CandidateSolution.

---

[2] There are exceptions to these generalizations that we postpone for the moment. For example, one can say of a known object 'It is blue', in which case that object (e.g., PAINTING-3) has a COREF link (e.g., to PAINTING-2) and is described by "COLOR blue". However, this pronominalization strategy only works for standalone utterances, not if 'the blue painting' is used as the case-role filler of another event, like 'I like the blue painting'.

Let us give just one more example of how a CandidateSolution fleshes out the corresponding CandidateSet. As we have already mentioned, there are many ways to ask someone to do something – ontologically, a REQUEST-ACTION event: *I would really appreciate it if you would X; it would be great <fantastic, terrific> if you would X; I'd like to ask you to (please) X; Would you be so kind as to X?*; and so on. Each such construction is anchored in the lexicon using a particular non-variable word. Table 4 shows the lexical sense that covers the first example above.

*Table 4.* Simplified lexical sense for appreciate-v8.

| | Top-level info | Syn-Struc | Sem-Struc |
|---|---|---|---|
| def | "indirect speech act 'I would (really) appreciate it if you (could/would) X" | syn-struc<br> use-example-binding t<br> n    $var1<br> aux  $var2 (root (will would))<br> adv  $var3 (root really) (opt +)<br> v    $var0<br> n    $var4 (root it)<br> prep $var5 (root if)<br> n    $var6 (root you)<br> aux  $var7 (root (could would)) (opt +)<br> v    $var8 | sem-struc<br> REQUEST-ACTION<br>   AGENT     ^$var1<br>   THEME    ^$var8<br>   BENEFICIARY ^$var6<br>   POLITENESS 1<br>   REFUSAL-OPPORTUNITY .5<br> [^$var2,3,4,5,7] null-sem+<br> example-bindings (I-1 would-2<br>   really-3 appreciate-0 it-4 if-5<br>   you-6 would-7 hit-8 the ball) |
| ex | "I would really appreciated it if you would make dinner." | | |

As we see from the sem-struc, there are only four main components to this meaning: the request, its agent, and its theme, and its beneficiary.[3] However, the request is expressed using a multi-word expression, all of whose component words – in the correct order and morphological form – must be introduced into the CandidateSolution at this stage.

## 3.3 Generate Candidate Sentences

Since we use an imported engine, SimpleNLG, to generate sentences from our CandidateSolutions, we point readers to Gatt & Reiter (2009) for its full description. We are not wed to this software package, and there are other options available. However, for our current experimentation, this is proving quite useful. Among the important features this engine provides us are generating the correct morphological forms of words based on features, generating the passive voice when our discourse analysis (e.g., comparison of new and known objects) suggests that it is the best option, and generating various forms of possessives.

## 3.4 Select the Best Sentence

Since we typically pass off multiple CandidateSolutions to SimpleNLG, we get multiple sentence realizations back. The next question is, How to choose the best one?

This seems like exactly the place where statistical methods might be useful. After all, statistically-trained language models – like the one underlying GPT-3 – are good at generating smooth, natural-sounding texts.[4] What we had *hoped* is that we could pass such a model a list of options for each of a sequence of sentences comprising a text and have it output the best sequence of sentences. This would be particularly helpful with respect to, e.g., selecting among synonyms that have no distinguishing features in our knowledge bases; realizing referring expressions; realizing the tense/aspect of verb forms; and capturing the topic-comment structure of discourse.

---

[3] The other elements of the sem-struc are property values (which do not get rendered directly by words) and indications of which words are not compositional (e.g., would, really, could, etc.).

A toy example will convey the point. The following is a natural sounding text: *Johnny jumped off the stairs onto his grandmother's couch. He heard the springs snap and realized he had broken it.* Consider just a handful of the sentence-realization options that could have been used to convey this meaning but would have sounded much worse: using Johnny's name more than once; saying 'the couch of his grandmother' rather than 'his grandmother's couch'; redundantly saying 'the springs of the couch' or 'the couch's springs' (though 'its springs' would have been fine); splitting the second sentence into two ('He heard… He realized'); using a passivization structure for the last clause ('he realized it has been broken by him'); getting the sequence of tenses wrong ('he realized he broke it'). Unfortunately, we were not able to find any off-the-shelf tools that would harness the power of GPT-3-style language models for our purposes here – they are just not set up as applications of this kind. Therefore, we need to rely wholly on knowledge-based selection criteria.

Although one might imagine that, given the significant history of work on NLG, there should be ready-made solutions for at least some component problems, this is – very unfortunately – not the case. As explained in McShane and Nirenburg (2021, Section 1.4.3), the theoretical and descriptive linguistic literature does not worry about computable prerequisites; the NLP literature has long focused on annotation-driven ML outside of cognitive systems; the NLP literature before the statistical turn offers descriptive and algorithmic food for thought but no actual solutions; and cognitive systems builders typically focus on narrow-domain applications whose requirements are quite different from those of developing robust language processing across domains.

At this early stage of our work on broad-coverage NLG, we do not have reportable results about how best to prune the candidate sentences but we do have ideas that will guide our ongoing work. For example, one branch of the linguistic theory called construction grammar focuses on frequency effects (e.g., Bybee, 2013). For NLG, frequency effects are particularly important in cases where language offers many paraphrases – as for requesting an action, discussed earlier. For NLG, the idea is for the agent to sound normal and avoid creativity and other flourishes that spice up human discourse but would sound bizarre coming from an artificial agent. (By contrast, we have spent significant time preparing our LEIAs to *understand* such inputs; McShane & Beale, Submitted.) For example, although the lexicon needs to account for such constructions as *I request that you X*, it is only used in hyper-formal contexts and should be avoided by agents. In principle, we could supplement the lexicon with features for all synonymous words and constructions that would reflect their frequency, but that would be a heavy knowledge engineering task. Instead, we are planning to automatically collect construction frequency using light parsing of the COCA (Davies, 2008-) corpus. We expect this to work reliably only for multi-component constructions, not single words, since most words are highly polysemous and frequency counts across their different meanings are unlikely to convey the selection preferences we are actually seeking.

---

[4] In ML-oriented, retrieval-based dialog systems, selecting the best candidate from available options is called response selection (Jia et al., 2020). However, it is entirely orthogonal to generating language from meaning representations.

Another inroad is our extensive past work on coreference in natural language understanding (cf. McShane and Nirenburg, 2021, Chapters 4-7). For NLU, we have found it useful to record constructions covering pairs of referring expressions – i.e., antecedents and their coreferents. The question is to what degree we can use those constructions to predict the best correlations of referring expressions during generation. Of course, there is a very large literature on referring expression selection in generation that will also inform this angle of our work. To give just a taste of the directions pursued: there has been extensive exploration of the Incremental Algorithm (Dale, 1989; Dale & Reiter, 1995) and its extensions, such as DIST-PIA (Williams & Sheutz, 2017); there are analyses of particular subproblems that might not immediately come to mind, such as the need to avoid false conversational implicatures through the choice of referring expressions (Reiter 1990); there are analyses of the deficiencies of proposed algorithms, including with respect to correlations with human studies (Deemter et al., 2012); and there are extensive linguistic studies in various schools of functional and discourse grammar (e.g., Keizer, 2014). One thing to note about much of this work, however, is that it reflects single-topic research thrusts: that is, it is relatively rare for the research effort to be exploring reference resolution along with discourse structure, surface realization, operating with a realistic-size lexicon that features extensive paraphrases, optimizing that lexicon to serve both NLU and NLG, and all the rest that we are trying to combine into a single, overarching, long-term, ever-advancing program of R&D.

## 4. Conclusions and Future Plans

This paper has explained how we are approaching NLG using the same knowledge bases, computational linguistic theory (ontological semantics), and agent architecture (OntoAgent) that have long supported our work on NLU. We are well into exploring linguistic eventualities and divvying up their treatment across multiple modules of NLG. We have an initial implementation of an NLG engine, called OntoGen, which can process simple inputs through the stage of generating candidate sentences using SimpleNLG.

In order to make work on broad-coverage NLG feasible, we have focused on its linguistic aspects, separated from content specification. We use the agent's NLU engine to provide content specifications as practice. Our goal is to create NLG capabilities that can be ported across domains and cognitive-system applications. Specifically, each cognitive system – which is likely to cover only a small domain – will provide content specifications that rely on its particular knowledge and reasoning. The our NLG engine will use generic, linguistically informed methods to generate the most appropriate English formulations.

Even when separated from content specification, there is nothing simple about NLG. However, this is not a reason to give up or despair. As we have learned from decades of work on NLU, it is counterproductive to view AI as all or nothing – there is plenty of room for results that are useful albeit short of human sophistication. Rules of thumb can go and far people can accommodate imperfect system output as long as it is understandable. It is our hope that readers will now better understand what NLG looks like from a knowledge-based perspective and will see that there are practical paths toward explainable, human-level language processing.

## Acknowledgments

This research was supported in part by Grant #N00014-19-1-2708 from the U.S. Office of Naval Research. Any opinions or findings expressed in this material are those of the authors and do not necessarily reflect the views of the Office of Naval Research.